# Multimodal rumor detection enhanced by external evidence and forgery features

**Han Li[1], Hua Sun[2,*]**

[1] Information Engineering School of Dalian Ocean University,Dalian Liaoning, China
[2] Information Engineering School of Dalian Ocean University,Dalian Liaoning, China
[2,*]Corresponding Author: Hua Sun. Email: sunhua@dlou.edu.cn

**ABSTRACT:** Social media increasingly disseminates information through mixed image–text posts, but rumors often exploit subtle inconsistencies and forged content, making detection based solely on post content difficult. Deep semantic-mismatch rumors, which superficially align images and texts, pose particular challenges and threaten online public opinion. Existing multimodal rumor detection methods improve cross-modal modeling but suffer from limited feature extraction, noisy alignment, and inflexible fusion strategies, while ignoring external factual evidence necessary for verifying complex rumors. To address these limitations, we propose a multimodal rumor detection model enhanced with external evidence and forgery features. The model uses a ResNet34 visual encoder, a BERT text encoder, and a forgery-feature module extracting frequency-domain traces and compression artifacts via Fourier transformation. While some existing approaches employ large-scale generative vision-language models for caption generation, their open-ended generation tendency produces verbose and stylistically inconsistent descriptions that introduce semantic noise and risk drifting from actual image content. To overcome this, we adopt BLIP—specifically pre-trained for vision-language alignment—which generates concise, image-faithful descriptions stylistically closer to news text, serving as a reliable semantic bridge across modalities. A BLIP-Driven Semantic Alignment Module jointly optimizes text–image and text–description contrastive losses, capturing inconsistencies at both visual and semantic levels. A gated adaptive feature-scaling fusion mechanism dynamically adjusts multimodal fusion and reduces redundancy. Experiments on Weibo and Twitter datasets demonstrate that our model outperforms mainstream baselines in M_Acc, recall, and F1 score.

**KEYWORDS:** Multimodal rumor detection; modality fusion; attention mechanism; image forgery

## 1 Introduction

The rapid development of social networks has enabled various types of information to be widely disseminated within an extremely short period of time. While this brings great convenience to users in accessing and sharing information, it also provides favorable conditions for the generation and spread of false content. Driven by the traffic economy, online rumors have significantly improved in terms of content breadth, dissemination speed, update frequency, and scope of influence, which not only disrupt the order of the online space but also erode social trust, posing a serious threat to public interests and social stability.

For example, in the special rectification campaigns recently carried out by relevant authorities, a large number of illegal accounts involved in rumors and massive amounts of rumor-related information have been handled in accordance with the law. These measures fully highlight the urgency and necessity of strengthening rumor governance in the social network environment. Meanwhile, with the continuous diversification of communication forms, posts on social platforms are no longer limited to text and often include images to enhance persuasiveness. Rumor makers frequently combine real or synthetic images with false text to form more deceptive image-text combinations (Tufchi et al.,2023) , which makes it more difficult for users to distinguish the authenticity of information and further increases the difficulty of rumor detection.

Therefore, authenticity detection for multimodal posts on social networks, aiming to achieve accurate rumor identification and curb the spread of false information, has become a current research hotspot in the field of information security and social network governance. Existing multimodal rumor detection models usually follow a consistent process: crawling posts from social platforms, extracting features of different modalities such as text and images, fusing these cross-modal features, and finally classifying rumors and non-rumors. Although researchers have developed various detection models through continuous exploration, there are still several common and prominent problems that limit the further improvement of detection performance.

In the stage of image modality feature extraction, most existing studies focus on extracting spatial domain features (Guo et al.,2023) but ignore the tampering traces hidden in the frequency domain (Wang et al., 2023). Since image tampering often leaves subtle clues in the frequency domain, the neglect of these features leads to insufficient ability of the model to identify visually tampered rumors. In terms of cross-modal alignment, although image-text consistency is widely used as a key judgment basis for rumor detection, the calculation method of this consistency and its specific role in model decision-making are often unclear (Wang et al.,2023), making it difficult to fully utilize the semantic correlation between text and images to achieve accurate rumor identification.

In terms of integrating social context features, due to the lack of explicit modeling of the interaction structure between posts, comments, and users, existing models are difficult to effectively capture the dissemination characteristics of rumors in social networks. In addition, structural noise is widespread in social network graphs, and traditional graph neural networks are prone to learning misleading context features when processing such graphs, which further affects the reliability of rumor detection results. In the feature extraction stage, most models rely on a single extraction method, resulting in insufficient mining of intra-modality heterogeneity and failure to fully capture the rich feature information contained in each modality.

In the feature fusion stage, some methods only use simple concatenation to combine cross-modal features, which cannot effectively handle the semantic differences and inconsistencies between different modalities. Although some studies introduce attention mechanisms to improve fusion performance, their usage methods and feature coverage are still limited (Li et al., 2023) , making it difficult to achieve deep fusion of cross-modal features and fully exert the complementary advantages of multiple modalities.

To address the above challenges and improve the accuracy and reliability of multimodal rumor detection, this paper proposes a multimodal rumor detection framework that integrates forgery feature extraction, a BLIP-Driven Semantic Alignment Module, and a gated fusion mechanism. Specifically, existing multimodal rumor detection methods mainly focus on semantic consistency while overlooking the authenticity of visual content, making them vulnerable to rumors containing manipulated images. Recent studies have attempted to incorporate frequency-domain forgery features, predominantly relying on the Discrete Cosine Transform (DCT) to characterize compression-related artifacts. However, such methods mainly emphasize block-level frequency characteristics associated with specific compression mechanisms, which may limit their ability to capture more diverse manipulation traces in complex social media scenarios. To address this issue, this paper introduces a forgery feature extraction module based on the Fourier transform during the visual feature extraction stage. By transforming images into the frequency domain, the proposed module obtains a global spectral representation that captures manipulation-induced spectral anomalies and subtle high-frequency artifacts that are difficult to perceive in the spatial domain. The extracted frequency-domain representations provide complementary authenticity cues beyond semantic features, enabling the model to better distinguish manipulated images from authentic ones and enhancing its robustness against visually manipulated rumors.

To solve the problem that traditional image-text alignment methods ignore the deep semantic consistency between modalities, this paper constructs a BLIP-Driven Semantic Alignment Module. Some existing methods leverage large-scale generative vision-language models to generate image captions as a means of bridging the modality gap. However, such models tend to produce verbose and stylistically diverse descriptions that deviate significantly from the concise, factual language characteristic of news text, inadvertently introducing semantic noise into the alignment process. Furthermore, their open-ended generation tendency may cause the produced captions to drift from the actual visual content, leading to misalignment between the generated description and the original image. To overcome these limitations, we employ BLIP, a model specifically pre-trained for vision-language alignment, which produces concise and image-faithful captions whose linguistic style is more consistent with news text. This design reduces the distributional discrepancy between generated descriptions and original news content, providing a more reliable semantic bridge for the subsequent contrastive learning objectives. The module includes text–image contrast and text–description contrast, jointly capturing semantic inconsistencies at both visual and semantic levels.

To identify rumors with superficial image-text consistency but deep semantic inconsistency, this paper introduces external evidence related to post content and uses a multi-head attention mechanism to fuse external evidence features with cross-modal features, enabling the model to more effectively detect rumors with deep inconsistency. Experimental results conducted on public multimodal rumor datasets Weibo and Twitter show that factors such as the number of external evidence and the introduction of forgery features have a significant impact on detection performance, and the proposed framework achieves better detection results compared with existing methods.

.

In summary, the main contributions of this paper are summarized as follows:

1. A multimodal rumor detection framework integrating forgery features and cross-modal semantic features is proposed. This framework can effectively cope with false text matching real images and visual tampering attacks, improving the model's adaptability to complex rumor forms.
2. A BLIP-Driven Semantic Alignment Module is proposed, which employs BLIP for alignment-oriented caption generation to reduce semantic noise and distributional discrepancy between visual and textual modalities, and jointly optimizes text–image and text–description contrastive losses to capture inconsistencies at both visual and semantic levels.
3. A gated modulation module is proposed, which couples dynamic gating with adaptive feature scaling. The gating mechanism realizes fine-grained dynamic fusion of text, image, and external evidence features to suppress redundancy, while the adaptive scaling mechanism provides coarse-grained magnitude constraint via a learnable global parameter, jointly enhancing the stability and discriminative power of the fused features.

## 2 Related Work

### *2.1 Monomodal Rumor Detection*

Research on rumor and fake-news detection in social media has formed a multidimensional technical system, evolving from traditional machine learning relying on handcrafted features, to deep-learning sequence modeling, graph-structure mining, multimodal fusion, and finally to recent advances leveraging large language models (LLMs) and reinforcement learning. Existing monomodal rumor detection methods utilize features extracted from either text or images to detect rumors. Early rumor detection studies mainly focused on mining prominent features in the rumor propagation process and applied traditional machine-learning classifiers for identification. Kwon et al. (Kwon et al., 2013) conducted an in-depth study on the temporal, structural, and linguistic characteristics of rumor propagation, proposing a periodic external shock model to capture the cyclical outbreak features of rumors over time. They also found that isolated nodes have a higher proportion in rumor propagation networks, and that the frequency of negation words in rumor content is significantly higher than in non-rumor content. These studies provided important feature foundations for subsequent deep-learning models. To overcome the limitations of handcrafted features, researchers began using models such as recurrent neural networks (RNNs) to automatically learn high-dimensional features. Ma et al. (Ma et al., 2016) first applied RNNs to capture temporal variation features of tweet content. On this basis, Chen et al. (Chen et al., 2018) proposed the dARNN model, which employs a GRU with a deep attention mechanism (a variant of recurrent neural networks) to capture long-range dependencies in post sequences, thereby enabling the model to filter out highly discriminative key clues from the highly redundant early information. In addition, Alkhodair et al. (Alkhodair et al., 2020) combined Word2Vec with LSTM to learn semantic features of text content, detecting sudden-news rumors related to emerging topics. Luvembe et al. (Luvembe et al., 2023a) proposed a deep-normalization-based attention mechanism to extract dual emotional features of rumors. Moreover, to achieve near-real time source blocking, Xu et al. (Xu et al., 2020) proposed a topic-driven rumor detection framework (TDRD), which uses LDA to extract the topic distribution vector of the source microblog and combines it with content features extracted by a CNN. The study verifies that the importance and ambiguity of topics are key factors in the emergence of rumors, thereby enabling early near-real-time detection that relies solely on the source post without depending on subsequent interactions. With a lot of multimedia messages, text information often cannot prevent complex fraud. Cao et al. (Cao et al., 2020) mentioned in the literature that visual information helps detect fake news. They divide the visual features into foreground, semantic, statistical and contextual. Building on this view, Qi et al. (Qi et al., 2019) A multi-domain visual neural network (MVNN) was proposed. The network has several parts: extracting manipulation artifacts using a subnet of the frequency domain subnet, capture of semantic content with the branches of the pixel domain and a combination of the focus of text and multimodal classification, emphasize the complementarity of the frequency space and the representation of the pixel space. In addition to the visual modeling, rumor detection has also begun leveraging large language models (LLM). Huang et al. (Huang & Sun, 2023) offer FakeGPT, which is used to assess the effectiveness of ChatGPT in the generation, interpretation and detection of rumors, and also to include the causal consciousness. To eliminate the shortcomings of large models in the comprehensive verification of the facts of Zhang and Gao (Zhang & Gao, 2023) offer the HiSS method. This method will destroy the claims that must be verified in small claims and then layered to verify them. This increases accuracy and leaves a clearer process of reasoning.

### *2.2 Multimodal Rumor Detection*

With the evolution of the social media ecosystem, the forms of rumor propagation have shifted from single-text to multimodal formats that include images and videos, which has driven researchers to leverage cross-modal complementarity to enhance detection performance. Early work mainly focused on multimodal feature extraction and fusion. Wang et al. (Wang et al., 2018) offered EANN. EANN can study the expression of text and images using event discriminators to reduce bias, specificity for events, and improve the effect. Nguyen et al. (Nguyen et al., 2019) modeled inter-article dependencies with a deep Markov random field to support classification. Kirchknopf et al. (Kirchknopf et al., 2021) further extended multimodal inputs by incorporating user comments and metadata via a multi-stream, hierarchical fusion design. Later-natured studies went beyond the “characteristic connection” and began to make obvious cross-modal interactions. Luvembe et al. (Luvembe et al., 2024) introduced CAF-ODNN, using complementary attention to capture fine-grained text–image associations and attenuate irrelevant semantic noise. Liu et al. (Liu et al., 2024) proposed MVACLNet, which promotes representation diversity through virtual augmented contrastive learning. Recently, the consistency-based approach as the main reference signal has become a semantic conflict of the image and text. Zhou et al. (Zhou et al., 2020) quantified cross-modal mismatch via similarity modeling, while Xue et al. (Xue et al., 2021) developed a framework that explicitly explores multimodal consistency patterns for detection. On the basis of capturing the semantic consistency between text and images, the model further introduces the SRM filter to extract the physical statistical features of images, and significantly improves the capability of identifying high-quality forged images by detecting physical tampering traces such as synthesis and modification. Abdelnabi et al. (Abdelnabi et al., 2022) proposed the CCN network, which targets “out-of-context” fake news by retrieving external evidence from the open web and performing multi-dimensional consistency checking using features extracted by CLIP. However, Zhang et al. (Zhang et al., 2023a) through large-scale data analysis, challenged the traditional assumption that image–text inconsistency implies falsehood, pointing out that rumor makers often deliberately select highly matched images to increase deception, providing a new perspective for similarity-based detection methods. Although the above methods have achieved significant progress in feature mining and consistency detection, most rely solely on the content of the posts themselves and lack support from external knowledge, resulting in misjudgments when facing carefully fabricated rumors. So, now we are beginning to study the "evidence-based reasoning," which is a new direction, but also the focus of this article. Wu et al. (Wu & Cao, 2024) two things were done: one was to create a Chinese multimodal rumor dataset called CSMRE, which contained evidence and the other to propose the EAMRD model. The model used a cross-attention mechanism to combine multimodal characteristics with the evidence found. To solve the problem that the model is subject to errors, Wu et al. (Wu et al., 2024a) developed the ERD-DC model. The model uses a two-channel approach: on the one hand, it quickly finds old rumors of evidence, and discovers new rumors with multidimensional functions on the other, which can effectively balance the efficiency and accuracy. In addition, Huang et al. (Huang et al., 2025) offered the model DEETSA. This model combines a double improvement of evidence with the perception of the similarity of the image text, and also provides effective integration of multi-source information through adaptive gated networks.

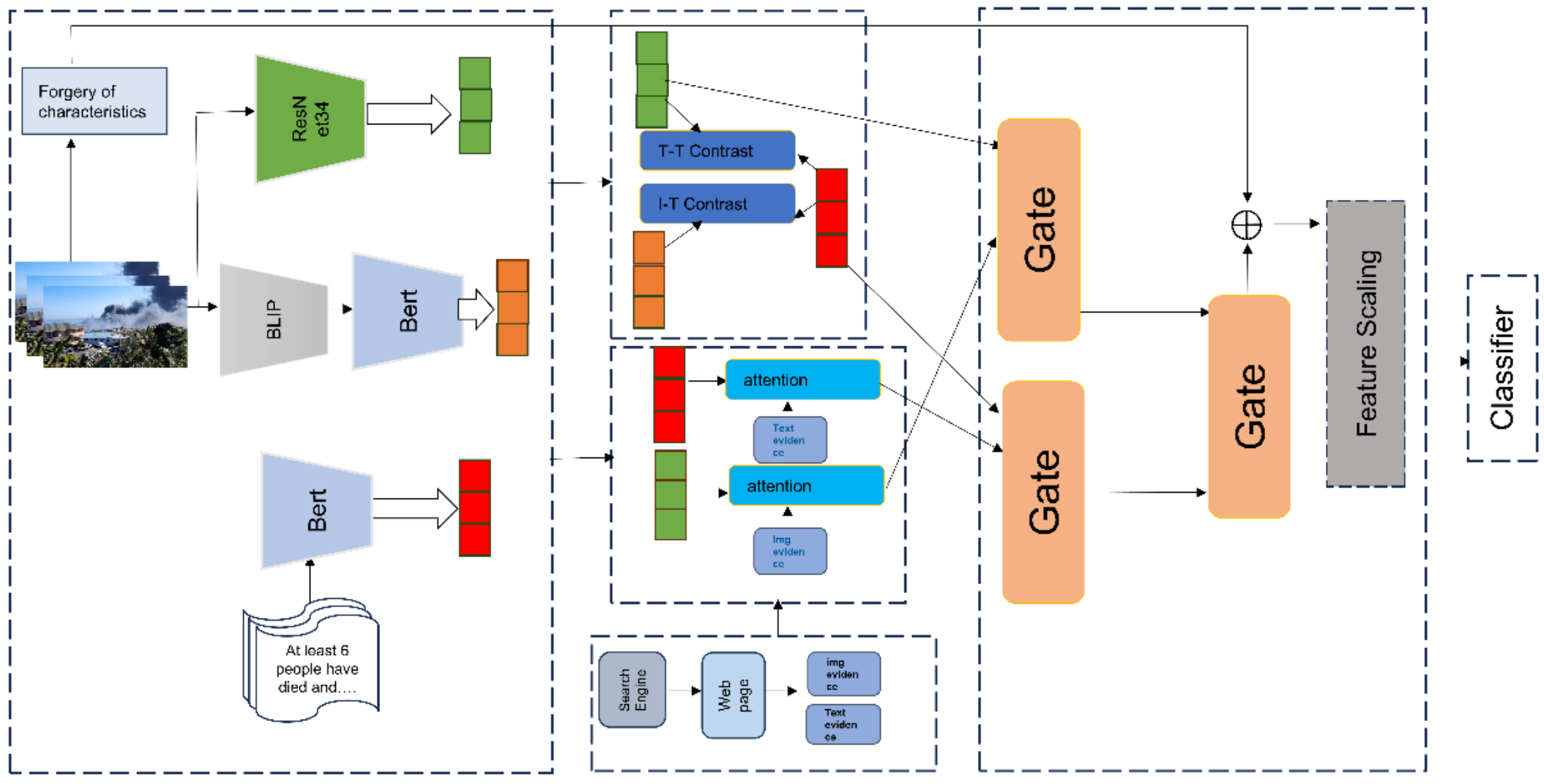

**Figure 1:** Model Structure Diagram.

## 3 Methodology

This paper presents a multimodal rumor detection framework integrating external evidence and digital forgery features through the interaction of feature encoding, BLIP-Driven Semantic Alignment Module, evidence fusion, and gated modulation modules. This multi-layered architecture enables the exhaustive extraction and synthesis of salient features from complex multimodal streams to optimize detection performance, with the comprehensive structural mechanisms detailed in Figure 1.

### *3.1 Feature Extraction*

In the multimodal rumor detection task, images and texts, as the two input modalities, differ not only in expression forms but also in information structure and semantic representation. To effectively extract meaningful features from these two modalities, the feature extraction module must be capable of deeply mining latent information in images and texts and mapping them into a unified feature space. This module combines image feature extraction, text feature encoding, and forgery-feature extraction, and enhances the alignment capability between images and texts through an effective fusion method.

*3.1.1 Text Features*

Text features are encoded using the BERT model. The BERT model, through a bidirectional Transformer encoder, is able to capture rich semantic information from context. After the textual data is input to BERT, the model learns representations of each word in context through its deep network. The goal of text feature encoding is to transform the original text into a high-dimensional semantic vector, which not only contains the specific meaning of words but also captures contextual relationships between words, thereby enabling learning of key information relevant to images, as follows:

$$h_T = BERT(T) \tag{1}$$

where $T$ represents the input textual information, $h_T \in R^{d_T}$ It represents the textual feature, and $d_T$ denotes the dimensionality of this feature.

*3.1.2 Image Features*

Image features are encoded using the ResNet34 network. ResNet34 is a deep residual network that addresses the gradient vanishing problem in deep neural networks through residual blocks, allowing the network to handle deeper structures. This model is capable of extracting high-dimensional visual features from images and learning key information contained in them, as follows:

$$h_I = ResNet34(I) \tag{2}$$

where $I$ denotes the input image, and the resulting representation $h_I \in R^{d_I}$ corresponds to its feature vector, with $d_I$indicating the dimensionality.

*3.1.3 Forgery Feature Extraction*

To further enhance the model's ability to recognize forgery content in images, this paper designs a forgery-feature extraction module based on Fourier transformation. This module extracts discriminative forgery features through frequency-domain transformation, multi-scale convolutional sampling, and global-dependency modeling.

Given a color image $X$ of size H×W, for each color channel, its two-dimensional discrete Fourier transform produces a complex spectrum $X_{comp}$. At the point (u,v), the computation is defined as:

$$X_{comp}(u,v) = \sum_{m=0}^{H-1} \sum_{n=0}^{W-1} X(m,n) \cdot e^{-j2\pi\left(\frac{um}{H}+\frac{vn}{W}\right)} \tag{3}$$

where $X(m,n)$is the pixel value at coordinate $(m,n)$in the spatial domain. $X_{\mathrm{comp}}(u,v)$ is the complex value in the frequency domain at coordinate $(u,v)$, representing the component of the wave with frequency $u$and $v$in the original image. According to Euler's formula $e^{j\theta} = \cos\theta + j\sin\theta (j^2 = -1)$, pixel values can be decomposed into a combination of sine and cosine waves.

A complex number contains a real part and an imaginary part, which cannot be directly processed by convolutional layers. Therefore, the amplitude spectrum $M(u,v)$ is calculated, representing the energy magnitude of each frequency component:

$$M(u,v) = | X_{comp}(u,v) | = \sqrt{Re\left(X_{comp}(u,v)\right)^2 + Im\left(X_{comp}(u,v)\right)^2} \tag{4}$$

where Re and Im denote the real and imaginary parts of the complex number, respectively.

Since the values of M(u,v) vary widely, directly inputting them into a neural network may cause gradient instability. Therefore, compression is applied:

$$X_{feat}(u,v) = ln(1 + M(u,v)) \tag{5}$$

To suppress random noise and retain only the mid- and high-frequency components, we define the low-frequency region $\Omega_{\text{low}}$ as a circular area at the center of the frequency spectrum, with the corresponding indicator function defined as:

$$I_{\Omega}(u,v) = \begin{cases} 1, & \sqrt{(u-H/2)^2 + (v-W/2)^2} \le R_{\text{low}} \\ 0, & \text{otherwise} \end{cases} \quad (6)$$

The low-frequency truncated feature map is then obtained by $X_{\text{freq}}(u,v) = X_{\text{feat}}(u,v) \cdot (1 - I_{\Omega}(u,v))$

Finally, the frequency-domain features of the three color channels are stacked to form $X_{\text{freq}} \in \mathbb{R}^{H\times W\times 3}$.

Then, $X_{\text{freq}}$ is input to a backbone network, and low-level features are extracted through convolutional layers:

$$F_{backbone} = CNN(X_{freq}) \quad (7)$$

where CNN denotes a convolutional neural network, and the output $F_{\text{backbone}}$ represents the initial image features.

Next, multi-scale features are obtained using different convolution kernels (1×1,3×3,5×5) and pooling operations, and the outputs are concatenated:

$$F_{ms} = [F_{branch1}, F_{branch3}, F_{branch5}, F_{branch_pool}] \quad (8)$$

To model long-range frequency dependencies across different image regions, the module introduces a multi-head self-attention mechanism. As shown in Figure 2 ,multi-head attention enhances model expressiveness by computing multiple parallel sets of distinct Query-Key-Value projections, enabling the simultaneous capture of diverse feature dependencies from multiple subspaces. $F_{\text{ms}}$ is reshaped and transposed into a sequence $Z \in R^{L\times C}$, where L=H'×W', C is the number of feature channels, and $H' \times W'$ is the height and width of the feature map.

For the $i$-th attention head, the computation is:

$$Attention(Q_i, K_i, V_i) = softmax\left(\frac{Q_i K_i^{\top}}{\sqrt{d_k}}\right) V_i \quad (9)$$

where $Q, K, V$ are the queries, keys, and values obtained from $Z$ via linear projection, and $\sqrt{d_k}$ is a scaling factor used to prevent large dot products from causing gradient vanishing.

By concatenating the outputs of $N$ heads and applying a linear projection, an enhanced feature sequence is obtained. Finally, global average pooling (GAP) followed by a fully connected (FC) layer generates the final forgery feature vector:

$$h_{forgery} = FC\left(GAP(MultiHead(Z))\right) \quad (10)$$

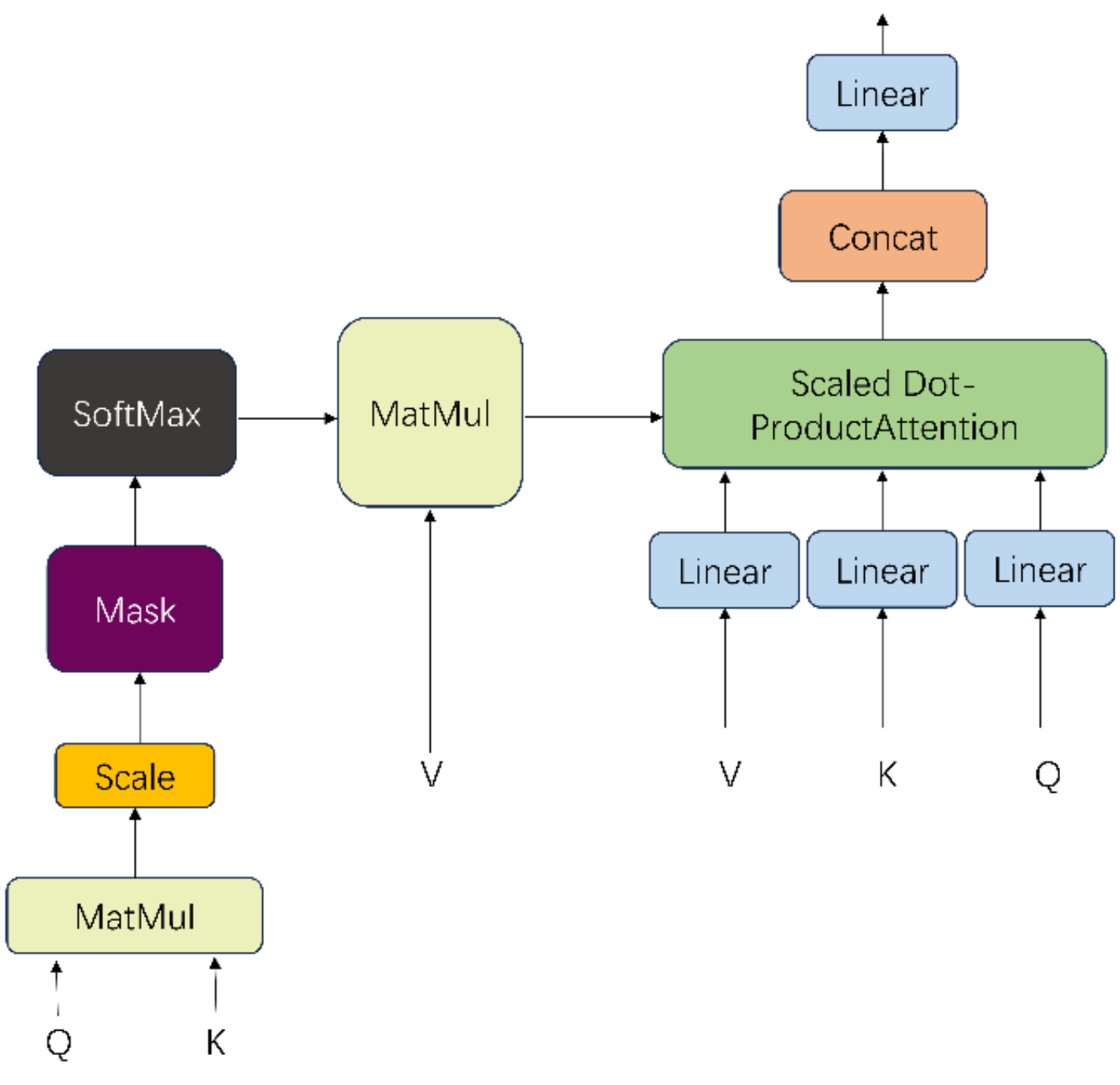

**Figure 2:** Multi-head attention.

### *3.2 Evidence Fusion Module*

To enhance semantic consistency across modalities through contrastive learning between text and evidence, as well as image and evidence, this paper introduces a multi-head attention-based evidence fusion module, inspired by the work of Huang et al. (Huang et al., 2025).

For textual evidence, we weight the evidence using a multi-head attention mechanism, allowing the selection of relevant evidence based on the main text features. Specifically, given the main text feature $h_T$ as the query, and the textual evidence $\tilde{E}_t$ as keys and values, multi-head attention is computed as follows. First, the main text feature $\mathrm{h_T}$ is mapped to a query vector, and the textual evidence features $\tilde{E}_t$, obtained from BERT, are mapped to key and value vectors:

$$Q = h_T W^Q, K = \tilde{E}_t W^K, V = \tilde{E}_t W^V \tag{11}$$

where $W^Q, W^K, W^V$ are learnable weight matrices.

To capture the relevance between the main text and evidence, the output of the $i$-th attention head $\mathrm{head}_i$ is calculated as:

$$head_i = Attention(Q_i, K_i, V_i) = softmax\left(\frac{Q_i K_i^{\mathrm{T}}}{d_k}\right) V_i \tag{12}$$

All head outputs are then concatenated and passed through a linear transformation to obtain the final attention-weighted evidence feature $\mathrm{H_a}$. The fused textual feature is denoted as $\mathrm{H_T^{'}}$:

$$H_T' = MultiHeadAttn\big(Q = h_T, K = \tilde{E}_t, V = \tilde{E}_t\big) \tag{13}$$

The processing of image evidence is similar to textual evidence. The image feature $h_I$ is used as the query, and the image evidence $\tilde{E}_i$, obtained by ResNet34, is used as keys and values:

$$H_I' = MultiHeadAttn(Q = h_I, K = \tilde{E}_i, V = \tilde{E}_i) \tag{14}$$

By integrating external evidence, the model can effectively align across multiple modalities and evidence, ensuring that it focuses on relevant evidence when processing multimodal information, thereby improving the accuracy of rumor detection.

### *3.3 BLIP-Driven Semantic Alignment Module*

To strengthen the model's capacity for detecting rumors that exhibit semantic inconsistencies between images and text, this paper designs a BLIP-Driven Semantic Alignment Module. Unlike existing methods that rely on large-scale generative vision-language models producing verbose and stylistically inconsistent captions, this module employs BLIP for alignment-oriented caption generation, reducing semantic noise and distributional discrepancy between visual and textual modalities. The module consists of two complementary contrastive learning objectives: semantic alignment between text and BLIP-generated image descriptions, and semantic alignment between text and visual image features. These two objectives constrain feature space consistency at the semantic and visual levels respectively, improving the model's comprehensive perception of text–image relationships and enabling effective detection of both surface-level and deep semantic mismatches between images and text.

### *3.3.1 Alignment-Oriented Image Caption Generation*

Unlike traditional methods that directly compute similarity in the visual feature space, This study employs the BLIP (Li et al., 2022) model to produce descriptive captions for images, thereby transforming visual inputs into interpretable textual semantic representations. Similar methods have also been used in (Huang et al., 2025) and achieved notable results.

Let the original image be $I$. After processing with the BLIP model, the corresponding textual description is $BLIP(I)$. The semantic feature $e_t$ is then extracted using BERT, as follows:

$$e_t = BERT(BLIP(I)) \tag{15}$$

### *3.3.2 Text–Image Description Semantic Alignment*

We first compute the semantic similarity between the original text feature $h_T$and the BLIP-generated image description feature $e_T$.

Let $u$and $v$denote the normalized text and image-description embeddings, respectively. For a batch of size $N$, we construct a similarity matrix $M_{TT}$, where each element $s_{i,j}$measures the similarity between the $i$-th text and the $j$-th image description::

$$s_{i,j} = u_i \cdot v_j^{\top} \tag{16}$$

To learn semantic consistency, we adopt a **contrastive learning objective** that encourages higher similarity for aligned pairs while suppressing mismatched pairs.

Importantly, **we adopt a label-aware positive pair construction strategy**. Specifically, only **non-rumor instances**, which exhibit reliable semantic consistency between text and image descriptions, are treated as positive pairs. Rumor instances are excluded from positive alignment due to potential cross-modal inconsistency.

For the $i$-th sample, the InfoNCE loss is defined as:

$$L_{TT} = -log \frac{exp(s_{i,i}/\tau)}{\sum_{j=1}^{N} exp(s_{i,j}/\tau)} \tag{17}$$

where $s_{i,i}$ is the score of the positive pair between the current text and its corresponding image description, $s_{i,j}(i \neq j)$ are negative pair scores, and $\tau$ is the temperature parameter used to adjust sensitivity to hard negatives.

This loss reinforces the consistency between image content and text descriptions at the semantic level by maximizing the similarity of matching samples (i=j) and minimizing the similarity of non-matching samples ($i \neq j$). It effectively detects semantic-level contradictions, for example, when the event described in text does not match the visual scene in the image.

### *3.3.3 Text–Image Visual Alignment*

To ensure semantic coherence at the visual level, the InfoNCE loss for the $i$-th sample is defined refer to Equation (18) and Equations (19):

$$L_{TI} = -log\frac{exp(s_{i,i}^{i}/\tau)}{\sum_{j=1}^{N} exp(\, s_{i,j}^{i}/\tau)} \tag{18}$$

$$s_{i,j}^{i} = u_i \cdot {g_j}^{T} \tag{19}$$

where $s_{i,j}^{i}$ represents the similarity between the $i$-th text and the $j$-th image, and $u$ and $g$ denote the normalized text and image features, respectively.

*3.3.4 Joint Contrastive Optimization*

The overall objective of the similarity perception module is:

$$L_{sim} = \lambda_{TT} L_{TT} + \lambda_{TI} L_{TI} \tag{20}$$

where $\lambda_{TT}$ and $\lambda_{TI}$ are balancing coefficients for semantic alignment and visual alignment, respectively.

Through dual-contrast constraints, the model simultaneously learns semantic-level alignment, ensuring consistency between textual descriptions and image content, and visual-level alignment, strengthening the correspondence between text and images in appearance and entities. This semantic–visual dual constraint allows the model to capture subtle cross-modal inconsistencies when encountering mismatched images and text or misleading headlines, thereby significantly improving detection accuracy.

***3.4 Gated Modulation Module***

In multimodal rumor detection, text and image semantics are highly complex and uncertain. Direct concatenation or weighted fusion easily causes semantic conflicts and feature redundancy. Traditional static methods fail to adaptively adjust inter-modal fusion ratios. Existing dynamic gating mechanisms, while enabling fine-grained proportion allocation, neglect coarse-grained control over the overall magnitude of fused features, resulting in dramatic fluctuations in the absolute scale of fused outputs across samples and leading to feature amplification and unstable training.

To address this, this paper proposes a gated modulation module that couples the gating mechanism with an adaptive feature-scaling mechanism, constructing a dual-level adaptive framework of fine-grained proportion regulation and coarse-grained magnitude constraint. The gating mechanism dynamically allocates the contribution ratio of each modality, while the adaptive scaling mechanism constrains the overall scale of fused features through a learnable global parameter.

*3.4.1 Gated Modulation Mechanism*

During the fusion stage, the model applies hierarchical gating control to text and image features to determine the contribution ratio of the main post information and external evidence. Specifically, let the main post text feature be $h_T, h_T \in R^{d_T}$, and the text feature enhanced by evidence attention be $\tilde{h}_T, \tilde{h}_T \in R^{d_T}$. Let the main post image feature be $h_I, h_I \in R^{d_I}$, and the image feature enhanced by evidence attention be $\tilde{h}_I, \tilde{h}_I \in R^{d_I}$.

The model first concatenates the two features and applies a linear transformation followed by a Sigmoid activation to obtain the inter-modal gating weights ,refer to Equation (21) and Equations (22):

$$\alpha_T = \sigma\big(W_T\big[h_T; \tilde{h}_T\big]\big), \tag{21}$$

$$\alpha_I = \sigma(W_I[h_I; \tilde{h}_I]). \tag{22}$$

where $\sigma(\cdot)$ denotes the Sigmoid function, and $W_T, W_I$ are learnable parameter matrices.

Then, the model performs weighted fusion of the main post and evidence features according to the gating coefficients, refer to Equation (23) and Equations (24).

$$h_T^{fusion} = \alpha_T \cdot h_T + (1 - \alpha_T) \cdot \tilde{h}_T, \tag{23}$$

$$h_I^{fusion} = \alpha_I \cdot h_I + (1 - \alpha_I) \cdot \tilde{h}_I. \quad (24)$$

The gating weights $\alpha_T$ and $\alpha_I$ reflect the importance of different modal features. When the main post information is sufficient, the gating factor tends to increase, allowing the model to retain more of the original semantics. When the external evidence contains strongly relevant or contradictory information, the gating coefficients decrease, enabling the model to actively incorporate external information for semantic correction.

### *3.4.2 Cross-Modal Gated Fusion*

After completing the single-modal gating, the model further fuses text and image semantic features in the cross-modal space. The gated text feature $h_T^{fusion}$ and image feature $h_I^{fusion}$ are mapped into the same latent space, refer to Equation (25) and Equations (26).

$$h_T' = W_C h_T^{fusion}, \quad (25)$$

$$h_I' = W_I h_I^{fusion}. \quad (26)$$

Then, a second-layer gating structure implements interaction fusion between the two modalities, refer to Equation (27) and Equations (28).

$$\beta = \sigma(W_{TI}[h_T'; h_I']), \quad (27)$$

$$h_{TI}^{fusion} = \beta \cdot h_T' + (1 - \beta) \cdot h_I'. \quad (28)$$

where $\beta$ denotes the cross-modal gating weight, controlling the proportion of text and image in the fused features.

Finally, forgery features are integrated to enhance the model's ability to recognize image manipulation, refer to Equation (29) and Equations (30).

$$h_{forgery} = W_S h_{forgery}, \quad (29)$$

$$h_{final} = [h_{TI}^{fusion}; h_{forgery}]. \quad (30)$$

where $[\cdot;\cdot]$denotes the feature concatenation operation.

### *3.4.3 Adaptive Feature Scaling Mechanism*

After obtaining the fused features, the model introduces an adaptive feature scaling mechanism to further improve robustness. This mechanism modulates the fused features through a learnable scaling parameter $\lambda$, refer to Equation (32) and Equations(33).

$$h_{final} = \lambda \cdot h_{final}, \quad (32)$$

$$\lambda \in (0,1]. \quad (33)$$

This gating parameter is jointly optimized with the other model parameters during training and is used to adaptively control the overall strength of the fused features. As a result, it mitigates noise interference caused by the accumulation of multimodal features and enhances the stability and generalization ability of the model during training.

## ***3.5 Classifier Design***

This study constructs a classification prediction module based on a two-layer perceptron, aiming to perform final category discrimination on the multimodal comprehensive feature $h_{\text{final}}$ obtained after gated fusion and adaptive feature scaling. This feature vector deeply integrates and enhances text semantics, image content, and forgery features, providing solid evidence for authenticity determination.

In implementation, the classifier first maps the fused features to a high-dimensional space through the first fully connected layer $\text{FC}_1$, embedding a LeakyReLU activation function to alleviate the "dead neuron" problem in deep networks and ensure continuous effective gradient flow during model iterations. The second fully connected layer $\text{FC}_2$ then maps the feature to a category prediction space of dimension $C$, outputting unnormalized logit vector $h_{\text{class}}$. Finally, the model applies the Softmax function to convert this vector into a predicted probability distribution $y$ refer to Equation (34) and Equations (35).

$$h_{class} = FC_2\left(LeakyReLU\left(FC_1\left(h_{final}\right)\right)\right), \tag{34}$$

$$y = Softmax(h_{class}). \tag{35}$$

where $y$ denotes the output probability distribution of the classifier.

The classifier is trained using the cross-entropy loss, which is commonly employed in classification tasks. The cross-entropy loss measures the difference between the predicted distribution and the ground-truth label distribution. For single-sample classification, the loss is defined as:

$$L_{class} = -\sum_{i=1}^{C} p_i \log(q_i) \tag{36}$$

where $C$ is the number of classes, $p_i$ is the ground-truth probability for class $i$, and $q_i$ is the predicted probability for class $i$.

## 4 Experiments

This section outlines the experimental protocol used to assess the proposed method. The MR2 datasets are introduced first, followed by implementation settings and evaluation metrics. Results are then reported for hyperparameter sensitivity, state-of-the-art comparisons, ablations, and error analysis. A final case study illustrates practical strengths and remaining limitations.

### *4.1 Datasets*

This study uses the MR2 dataset, released by Hu et al. (Hu et al., 2023), which is specifically designed for multimodal rumor detection. MR2 consists of two sub-datasets from Twitter and Weibo, covering content from English and Chinese social media platforms. Each sample contains a text paragraph and an accompanying image. The samples are labeled into three categories: rumor, non-rumor, and unconfirmed rumor.

The construction of the MR2 dataset not only involves collecting text–image pairs but also retrieving additional textual and visual evidence from the web. Image evidence is typically obtained through reverse image search of the original image, while textual evidence is acquired by searching the original text to collect relevant web page titles, summaries, and other textual information.

### *4.2 Experimental Settings and Evaluation Metrics*

In this experiment, the dataset is split into training, validation, and test sets with a ratio of 8:1:1. The maximum text length is set to 40 tokens to accommodate most Twitter and Weibo samples. In practice, the maximum number of text and visual evidence items for each post is set to 5, and 8 attention heads are used for cross-attention.

The classification task employs a two-layer perceptron network. During training, the model uses the AdamW optimizer with an initial learning rate of 0.00006, a batch size of 32, and a total of 8 training epochs. Learning rate adjustment is performed using the ExponentialLR scheduler to dynamically tune the learning rate and improve training efficiency.

To prevent overfitting, an early stopping strategy with a patience of 2 is employed, meaning training stops when validation performance no longer improves.

We utilize Mean Class Accuracy (M_Acc), Precision (Pre), Recall (Rec), and F1-score (F1) for each class as performance evaluation metrics. The calculations are as follows:

$$M_Acc = \frac{1}{C}\sum_{c=1}^{C} Acc_c \tag{37}$$

$$Acc_c = \frac{TP_c}{N_c} \tag{38}$$

$$Pre_c = \frac{TP_c}{TP_c + FP_c} \quad (39)$$

$$Rec_c = \frac{TP_c}{TP_c + FN_c} \quad (40)$$

$$F1_c = \frac{2 \cdot Pre_c \cdot Rec_c}{Pre_c + Rec_c} \quad (41)$$

where $c$denotes the class index and $C$denotes the number of classes. $N_c$represents the total number of samples in class $c$. $TP_c$, $TN_c$, $FP_c$, and $FN_c$denote the number of true positives, true negatives, false positives, and false negatives for class $c$, respectively.

### *4.3 Experimental Results and Analysis*

#### *4.3.1 Comparison with Baseline Methods*

To verify the effectiveness of our proposed method, we selected the following six methods as comparison baselines:

CLIP (Radford et al., 2021): Uses the image encoder of CLIP to extract image features, and classifies the feature vector corresponding to the special token "CLS" via a linear layer.

LSTM_word2vec (Alkhodair et al., 2020): A text-only detection method. It employs word2vec and LSTM to extract semantic features from text content for rumor detection.

DEDA (Luvembe et al., 2023a): A text-only detection method. It utilizes a deep normalized attention mechanism to extract sentiment features from text for rumor judgment.

MVACLNet: (Liu et al., 2024): A multimodal rumor detection method that utilizes virtual augmentation contrastive learning and hierarchical textual feature extraction to enhance cross-modal representation learning and improve rumor detection performance.

CCN (Abdelnabi et al., 2022): A multimodal detection method specifically designed to address the selective distortion problem in text–image mismatches. It uses a Consistency Checking Network (CCN) to learn and determine whether external evidence is consistent with post content.

MRAN (Yang et al., 2024): A multimodal method that detects fake news through an attention network modeling both intramodal and intermodal relationships.

DEETSA(Huang et al., 2025): A multimodal detection method that enhances judgment by retrieving external textual and visual evidence, and quantifies the similarity between text and images to identify semantic inconsistency in rumors.

To verify the effectiveness of the method, this paper compares six rumor detection methods on two datasets of different languages, Weibo and Twitter. Table 1and 2 show that the method in this paper achieves the best results in both cases. It reaches a M_Acc of 94.9% on the Weibo dataset and 94.1% on the Twitter dataset. These results verify the effectiveness of the multimodal fusion strategy proposed in this paper and demonstrate its leading performance in cross-linguistic scenarios, indicating that the method possesses good language-independent characteristics.

The results show that, due to ignoring information in either text or graphic features, rumor detection methods relying on a single modality are generally inferior to multimodal rumor detection methods. Multimodal rumor detection methods can distinguish rumors through more information; therefore, the M_Acc of multimodal models is significantly improved. Among many multimodal baseline models, this method is slightly superior to DEETSA, which introduces external evidence, and MRAN, which introduces a cross-attention mechanism; the advantage lies in the addition of image forgery feature enhancement, demonstrating the importance of forged image identification for rumor detection.

**Table 1:** The comparative experiment on the Weibo dataset.

| | Non-rumor | Rumor | Unverified |
|---|---|---|---|

| | M_Acc | Precision | Recall | F1 Score | Precision | Recall | F1 Score | Precision | Recall | F1 Score |
|---|---|---|---|---|---|---|---|---|---|---|
| CLIP | 82.8 | 88.7 | 90.3 | 89.5 | 87.2 | 75.6 | 81.0 | 78.2 | 82.5 | 80.3 |
| DEDA | 88.6 | 93.6 | 93.4 | 93.5 | 84.6 | 81.5 | 83.0 | 89.1 | 90.9 | 90.0 |
| MVACLNet | 89.2 | 95.0 | 95.0 | 95.0 | 91.8 | 75.8 | 83.1 | 88.7 | 96.6 | 92.5 |
| CCN | 92.3 | 96.7 | 95.5 | 96.1 | 85.7 | 87.9 | 86.8 | 93.9 | 93.5 | 93.7 |
| MRAN | 92.7 | 97.1 | 94.4 | 95.7 | 84.4 | 91.2 | 87.7 | 94.5 | 92.4 | 93.4 |
| DEETSA | 93.7 | 94.3 | 95.9 | 95.1 | 92.6 | 91.0 | 91.8 | 94.6 | 94.3 | 94.4 |
| ours | 94.9 | 95.1 | 96.3 | 95.7 | 88.6 | 96.1 | 92.2 | 97.3 | 92.3 | 94.7 |

**Table 2:** The comparative experiment on the Twitter dataset.

| | | Non-rumor | | | Rumor | | | Unverified | | |
|---|---|---|---|---|---|---|---|---|---|---|
| | **M_Acc** | **Precision** | **Recall** | **F1 Score** | **Precision** | **Recall** | **F1 Score** | **Precision** | **Recall** | **F1 Score** |
| CLIP | 83.8 | 92.6 | 85.3 | 88.8 | 81.3 | 79.8 | 80.5 | 81.1 | 86.3 | 83.6 |
| DEDA | 88.8 | 91.9 | 93.4 | 92.6 | 84.9 | 79.5 | 82.1 | 91.8 | 93.4 | 92.6 |
| MVACLNet | 91.0 | 95.5 | 96.0 | 95.8 | 85.4 | 81.4 | 83.3 | 94.1 | 95.7 | 94.9 |
| CCN | 91.8 | 94.8 | 95.7 | 95.2 | 88.7 | 84.5 | 86.5 | 93.8 | 95.2 | 94.5 |
| MRAN | 92.5 | 93.3 | 97.1 | 95.2 | 93.3 | 86.0 | 89.5 | 93.7 | 94.3 | 94.0 |
| DEETSA | 93.8 | 95.6 | 98.5 | 97.0 | 85.8 | 89.1 | 87.5 | 97.4 | 93.6 | 95.5 |
| ours | 94.1 | 97.9 | 95.5 | 96.7 | 85.5 | 91.5 | 88.4 | 96.8 | 95.4 | 96.1 |

To provide a more intuitive comparison between the models, this paper presents the results in the form of a bar chart. It can be observed that the precision, recall, and F1-score for all three categories show an upward trend, as illustrated in Figure 3.

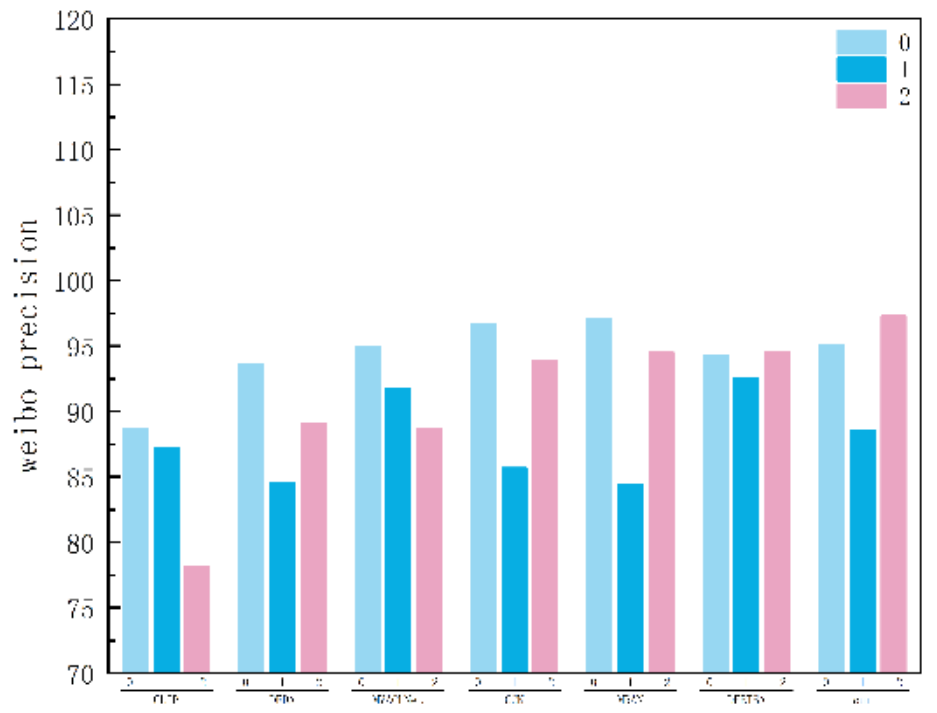

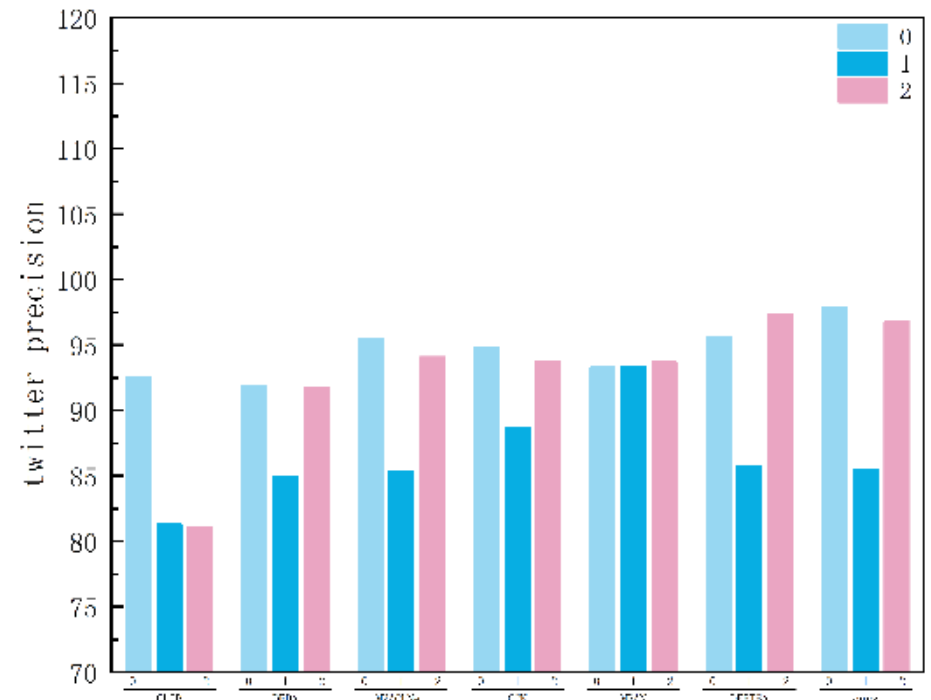

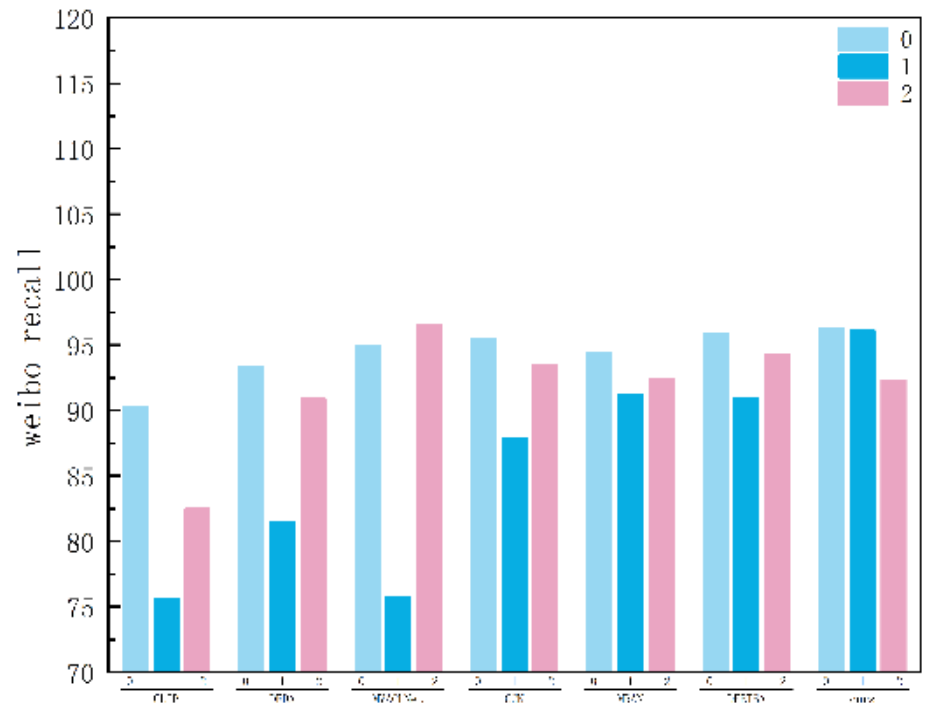

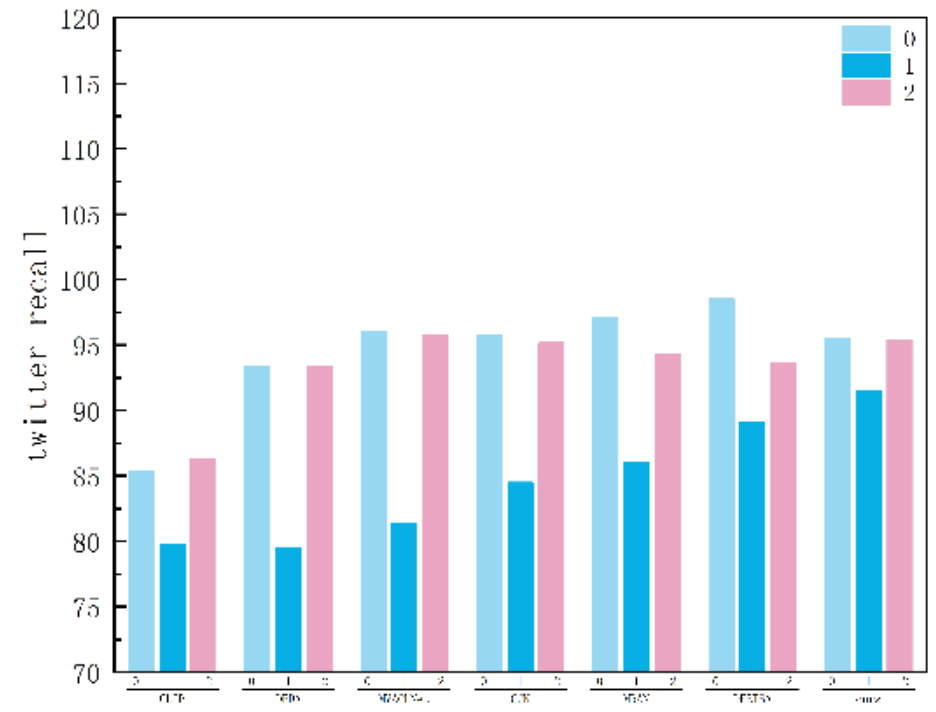

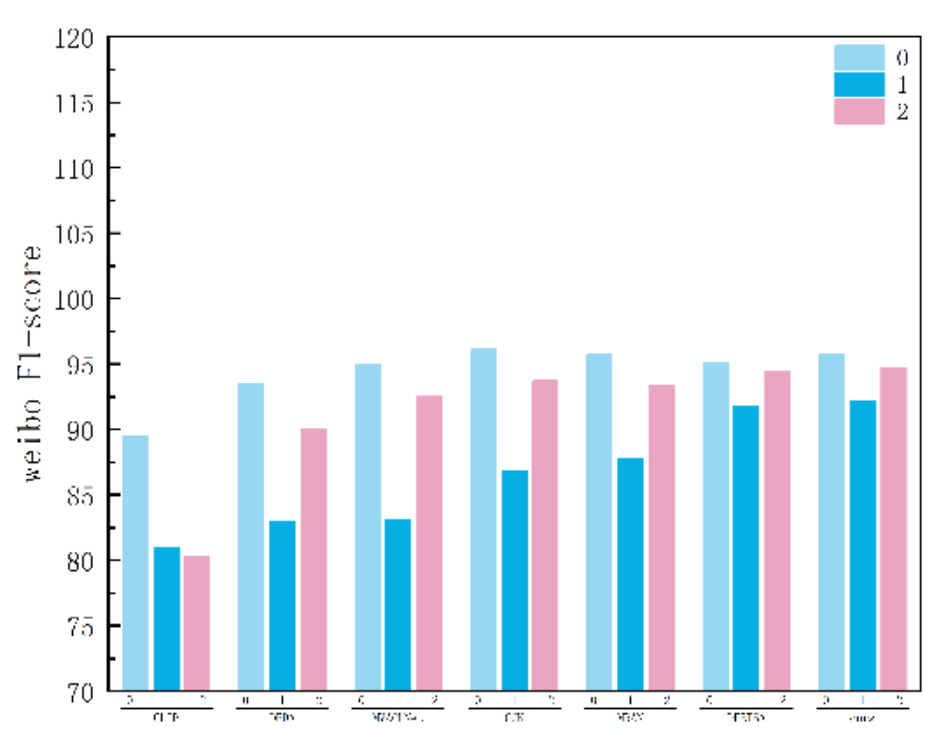

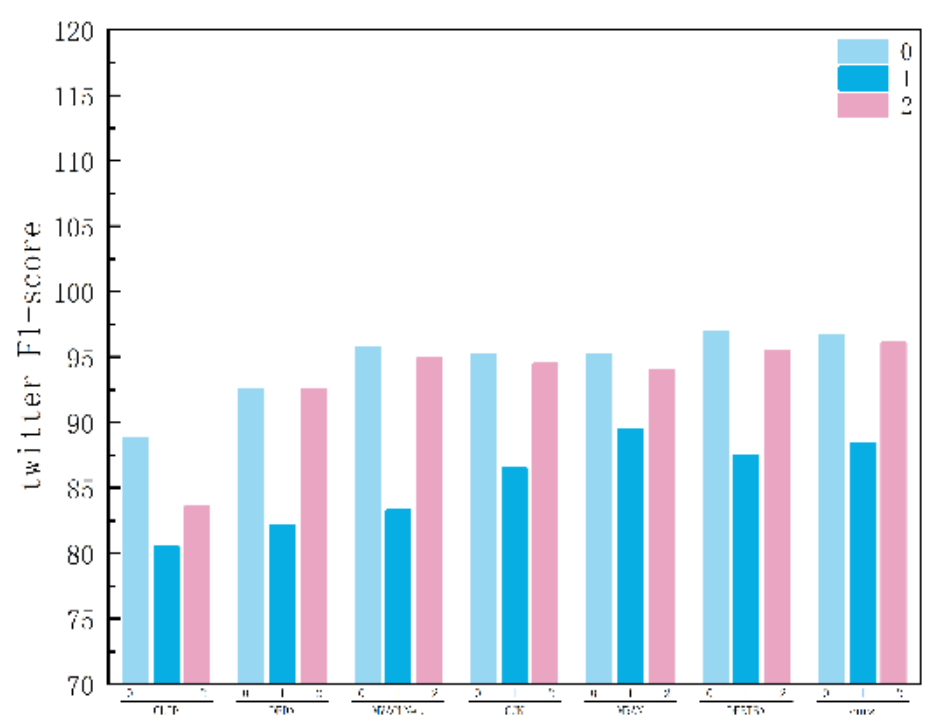

**Figure 3:** Comparative Experimental Bar Chart.

*4.3.2 Ablation Study*

To verify the contribution of each module, we design the following seven ablation experiments:

w/o D C: The BLIP-Driven Semantic Alignment Module is removed.

w/o Evidence: Both textual and visual evidence are removed.

w/o F S: Only the adaptive feature scaling mechanism is removed.

w/o Forgery: The forgery feature extraction module is removed.

w/o Gating: The gated fusion and adaptive feature scaling mechanisms are removed.

The results on the Weibo and Twitter datasets are shown in Tables 3 and 4, respectively. By comparing the full model with these seven variants, the ablation study effectively validates the necessity and effectiveness of each component in the model.

Table 3 and 4 show the results on the Weibo and Twitter datasets, respectively. It can be seen that the complete model achieves optimal performance across most metrics. This reflects the important collaborative effects between the various modules. In the experiments, removing the gating module and the image forgery feature extraction module caused the most significant performance degradation. After removing the gating mechanism, the M_Acc on the Weibo dataset dropped by 3.5%, and the impact on the F1 score of the rumor category was significant. This proves the effectiveness of jointly optimizing intra-modal and inter-modal semantic consistency. After removing the image forgery feature extraction, the M_Acc on the Twitter dataset decreased by 3.0%. This indicates that capturing image tampering traces is crucial for rumor detection, especially on image-rich platforms like Twitter. Removing the evidence fusion module also caused a clear decline in performance, illustrating that external evidence plays a major auxiliary role in rumor detection.

Compared with the w/o FS variant, the complete model achieves improvement in M_Acc, indicating a more balanced classification performance across different classes. This demonstrates that the adaptive feature scaling module effectively calibrates the fused feature representation, enabling the model to learn more discriminative and robust multimodal features.

In addition, when the BLIP-Driven Semantic Alignment Module is removed, the model performance also drops noticeably, which demonstrates that aligning image-generated semantic descriptions with textual content is essential for capturing fine-grained cross-modal semantic consistency and improving detection robustness.

**Table 3:** The ablation experiment results on the Weibo dataset.

| | | **Non-rumor** | | | **Rumor** | | | **Unverified** | | |
|---|---|---|---|---|---|---|---|---|---|---|
| | **M_Acc** | **Precision** | **Recall** | **F1 Score** | **Precision** | **Recall** | **F1 Score** | **Precision** | **Recall** | **F1 Score** |
| w/o DC | 93.1 | 96.7 | 97.1 | 96.9 | 92.8 | 87.1 | 89.9 | 92.5 | 95.1 | 93.8 |
| w/o Evidence | 91.9 | 98.3 | 92.9 | 95.5 | 88.5 | 86.5 | 87.5 | 91.8 | 96.3 | 94.0 |
| w/o FS | 93.5 | 97.1 | 95.9 | 96.5 | 90.0 | 89.3 | 89.6 | 94.2 | 95.4 | 94.8 |
| w/o Forgery | 92.2 | 96.7 | 97.9 | 97.3 | 87.9 | 85.9 | 86.9 | 92.6 | 92.9 | 92.8 |
| w/o Gating | 91.4 | 95.1. | 95.5 | 95.3 | 90.9 | 83.7 | 87.2 | 91.7 | 95.1 | 93.4 |
| ours | 94.9 | 95.1 | 96.3 | 95.7 | 88.6 | 96.1 | 92.2 | 97.3 | 92.3 | 94.7 |

**Table 4:** The ablation experiment results on the Twitter dataset..

| | | **Non-rumor** | | | **Rumor** | | | **Unverified** | | |
|---|---|---|---|---|---|---|---|---|---|---|
| | **M_Acc** | **Precision** | **Recall** | **F1 Score** | **Precision** | **Recall** | **F1 Score** | **Precision** | **Recall** | **F1 Score** |
| w/o DC | 92.4 | 93.8 | 98.5 | 96.1 | 87.2 | 84.5 | 85.8 | 96.4 | 94.3 | 95.3 |
| w/o Evidence | 91.8 | 96.4 | 94.0 | 95.2 | 80.7 | 87.6 | 84.0 | 96.0 | 93.9 | 94.9 |
| w/o FS | 92.9 | 97.0 | 98.0 | 97.5 | 93.0 | 82.9 | 87.7 | 93.9 | 97.9 | 95.9 |
| w/o Forgery | 91.1 | 92.9 | 98.5 | 95.6 | 93.5 | 77.5 | 84.8 | 93.9 | 97.2 | 95.5 |
| w/o Gating | 91.7 | 95.5 | 95.5 | 95.5 | 82.7 | 85.3 | 84.0 | 95.7 | 94.3 | 95.0 |
| ours | 94.1 | 97.9 | 95.5 | 96.7 | 85.5 | 91.5 | 88.4 | 96.8 | 95.4 | 96.1 |

### *4.3.3 t-SNE Feature Visualization Analysis*

Figure 4 show the visualization results of feature fusion on the Weibo and Twitter datasets. Blue, brown, and cyan represent non-rumor, rumor, and unverified samples, respectively. It can be seen that different samples form distinct clusters.

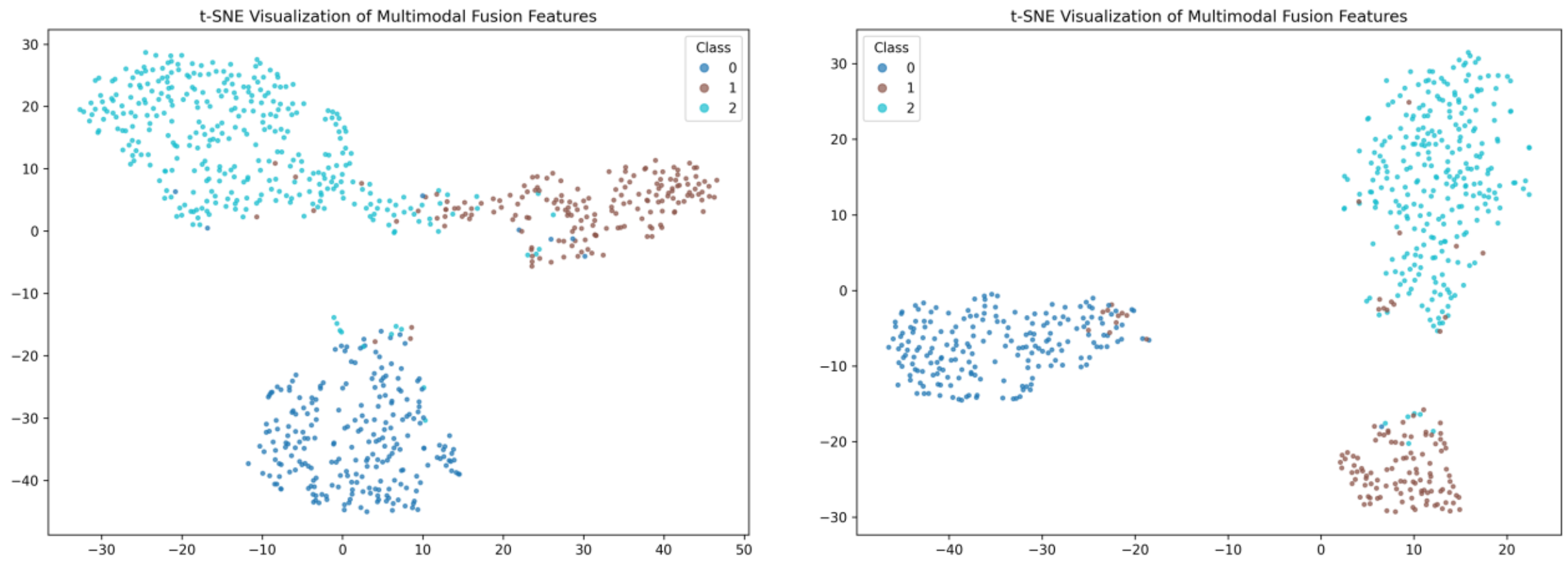

**Figure 4:** t-SNE Feature Visualization..

*4.3.4 Effect of Evidence Quantity*

The maximum number of retrieved evidence items is a critical hyperparameter: too little evidence limits context for verification, whereas too much increases the risk of irrelevant or incorrect matches. Figure 5 reports a sweep of retrieved text/visual evidence from 1 to 9 items and the corresponding M_Acc on Weibo and Twitter. Performance peaks at 5 evidence items on both datasets, indicating that evidence quantity and quality jointly shape verification: under-retrieval leaves claims under-supported, while over-retrieval injects noise that degrades classification.

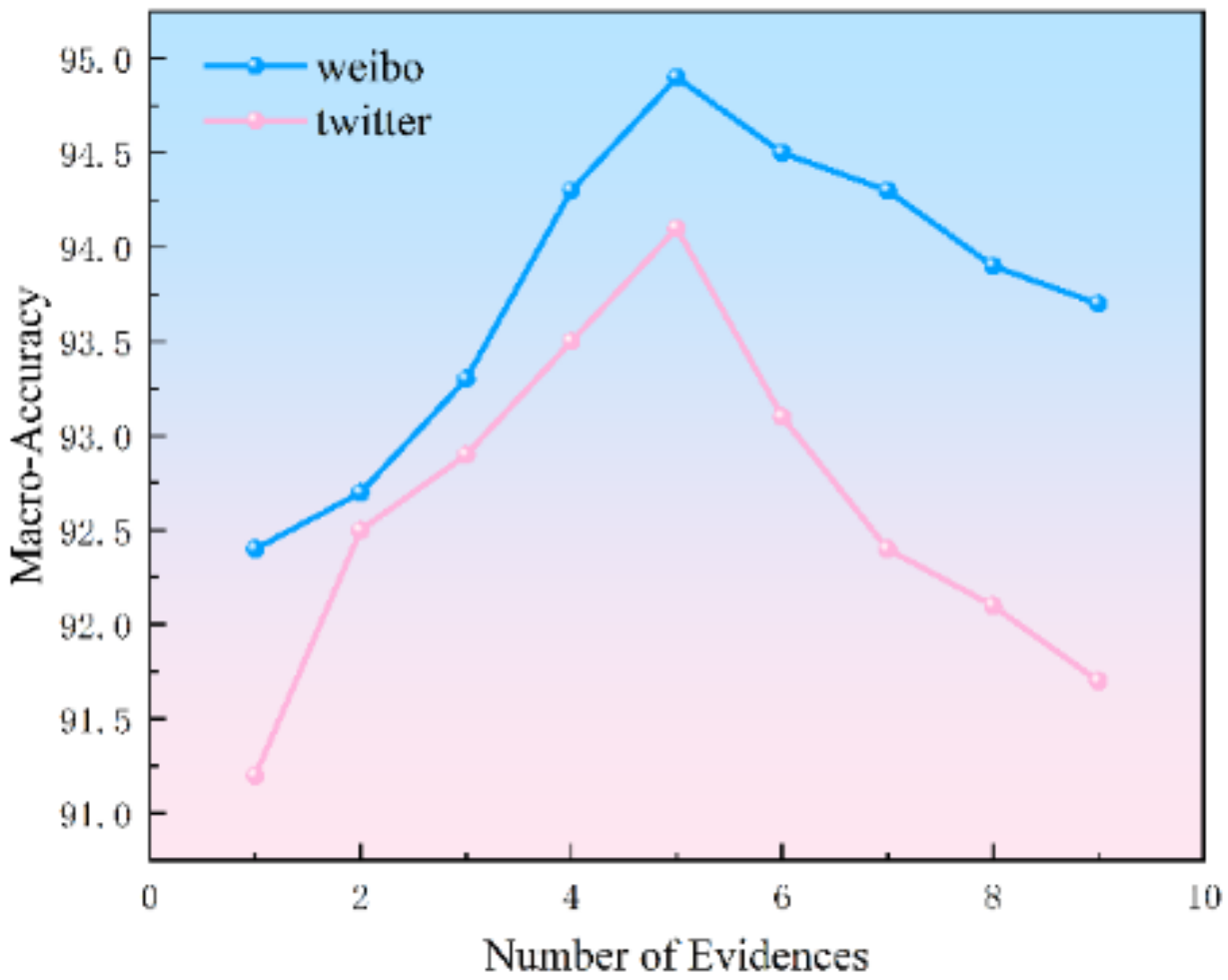

**Figure 5:** The impact of evidence quantity..

*4.3.5 Error Analysis*

To conduct a detailed analysis of prediction errors, we visualized the confusion matrices of the test results, as shown in Figure 6, where the horizontal and vertical axes represent the predicted labels and ground truth labels, respectively. Although the model's F1 score exceeds 94% in the non-rumor and unverified categories, the rumor identification performance on Twitter drops to 88.4%, while on Weibo it is 92.2%. This highlights the difficulty of identifying misinformation on Twitter. The precision and recall on the Weibo dataset indicate that the model tends to identify information as rumors. In the unverified category, the model exhibits high precision and relatively low recall, which indicates that the model is very cautious when confirming this type of information.

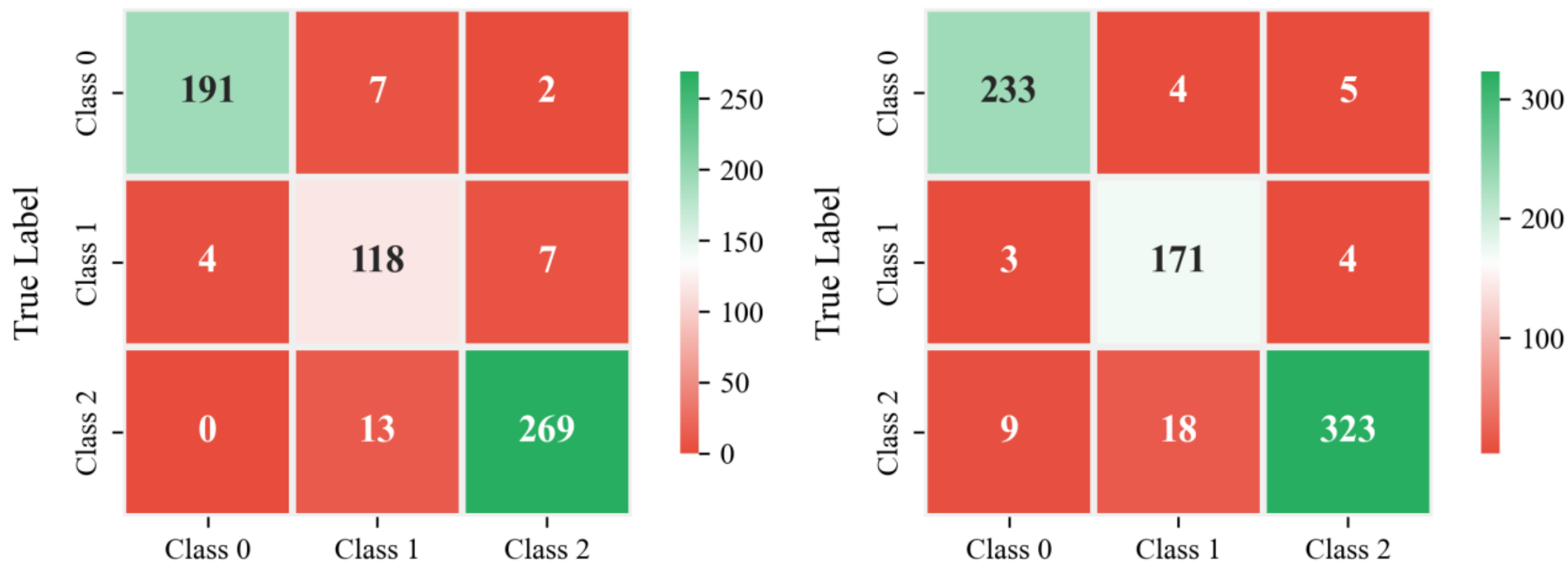

**Figure 6:** Confusion matrix visualization**.**

## 5 Conclusion

This paper proposes a multimodal rumor detection framework integrating external evidence and forgery features, which significantly improves detection performance through a BLIP-Driven Semantic Alignment Module and a gated modulation module. The BLIP-Driven Semantic Alignment Module employs BLIP for alignment-oriented caption generation, reducing semantic noise and modality gap, and jointly optimizes text–image and text–description contrastive losses to capture inconsistencies at both visual and semantic levels. Experiments on the Weibo and Twitter datasets demonstrate improvements in multimodal reasoning, image forgery sensitivity, and cross-modal semantic alignment. The research results indicate that a balance exists between evidence and noise: insufficient evidence lacks context, while excessive evidence hinders the integration of relevant content; therefore, the quantity of evidence must be carefully selected. By introducing a feature fusion method with an adaptive gated mechanism, the model can automatically adjust fusion weights based on feature importance, effectively reducing the negative impact of noise on performance. Future work will explore more fine-grained integration methods and extend evidence collection strategies in high-noise and complex scenarios.